\documentclass[sigconf]{acmart}

\AtBeginDocument{%
  }

\copyrightyear{2024}
\acmYear{2024}
\setcopyright{acmlicensed}
\acmConference[CIKM '24] {Proceedings of the 33rd ACM International Conference on Information and Knowledge Management}{October 21--25, 2024}{Boise, ID, USA.}
\acmBooktitle{Proceedings of the 33rd ACM International Conference on Information and Knowledge Management (CIKM '24), October 21--25, 2024, Boise, ID, USA}
\acmISBN{979-8-4007-0436-9/24/10}
\acmDOI{10.1145/3627673.3680007}

\usepackage{caption}
\usepackage{subcaption}
\usepackage[ruled,linesnumbered]{algorithm2e}
\usepackage{balance}
\makeatletter
\newcommand{\linebreakand}{%
  \end{@IEEEauthorhalign}
  \hfill\mbox{}\par
  \mbox{}\hfill\begin{@IEEEauthorhalign}
}
\usepackage{adjustbox}
\usepackage{hyperref}
\makeatother

\begin{document}
\title{You Can't Ignore Either: Unifying Structure and Feature Denoising for Robust Graph Learning}


\author{Tianmeng Yang}
\authornote{Both authors contributed equally to this research.}
\affiliation{
\institution{Peking University}
\city{Beijing}
\country{China}
}
\email{youngtimmy@pku.edu.cn}

\author{Jiahao Meng}
\authornotemark[1]
\affiliation{
\institution{Peking University}
\city{Beijing}
\country{China}
}
\email{mengjiahao@stu.pku.edu.cn}

\author{Min Zhou}
\affiliation{
\institution{Huawei Cloud}
\city{Shenzhen}
\country{China}
}
\email{zhoumin27@huawei.com}

\author{Yaming Yang}
\affiliation{
\institution{Peking University}
\city{Beijing}
\country{China}
}
\email{yamingyang@stu.pku.edu.cn}

\author{Yujing Wang}
\author{Xiangtai Li}
\affiliation{
\institution{Peking University}
\city{Beijing}
\country{China}
}
\email{{yujwang, lxtpku}@pku.edu.cn}

\author{Yunhai Tong}
\affiliation{
\institution{Peking University}
\city{Beijing}
\country{China}
}
\email{yhtong@pku.edu.cn}


\renewcommand{\shortauthors}{Tianmeng Yang et al.}

\begin{abstract}
Recent research on the robustness of Graph Neural Networks (GNNs) under noises or attacks has attracted great attention due to its importance in real-world applications. Most previous methods explore a single noise source, recovering corrupt node embedding by reliable structures bias or developing structure learning with reliable node features. However, the noises and attacks may come from both structures and features in graphs, making the graph denoising a dilemma and challenging problem. In this paper, we develop a unified graph denoising (UGD) framework to unravel the deadlock between structure and feature denoising. Specifically, a high-order neighborhood proximity evaluation method is proposed to recognize noisy edges, considering features may be perturbed simultaneously. Moreover, we propose to refine noisy features with reconstruction based on a graph auto-encoder. An iterative updating algorithm is further designed to optimize the framework and acquire a clean graph, thus enabling robust graph learning for downstream tasks. Our UGD framework is self-supervised and can be easily implemented as a plug-and-play module. We carry out extensive experiments, which proves the effectiveness and advantages of our method. Code is avalaible at \href{https://github.com/YoungTimmy/UGD}{https://github.com/YoungTimmy/UGD}.
\end{abstract}




\maketitle

\section{Introduction} 
By extending deep learning on graphs, graph neural networks have achieved great success and been applied to many important scenarios, such as social networks~\cite{kumar2022influence,song2023xgcn}, recommendation~\cite{ying2018graph,wu2022graph_rec_survey} and financial transaction~\cite{dou2020care-gnn,yang2023mitigating}. Most GNNs~\cite{kipf2016semi,velivckovic2017graph,hamilton2017inductive} follow a message passing framework~\cite{gilmer2017neural} to learn node representations for downstream tasks, relying on a favorable neighborhood aggregation including both structure connections and node features.

Unfortunately, real-world graphs are often noisy or incomplete due to error-prone data measurement or collection. For example, in a user-item graph, users may have imperfect profiles (feature missing) or misclick some unwanted items (error link)~\cite{o2006detecting}. Malicious accounts may camouflage themselves, including adjusting their behaviors (feature camouflage) and connecting with several benign entities (structure camouflage)~\cite{dou2020care-gnn}. What is worse, recent studies show that GNNs are vulnerable to adversarial attacks~\cite{dai2018adversarial,jin2020adversarial_survey}. Disturbing graph structure or node features would greatly degrade the performance of GNNs and may lead to severe consequences for critical applications requiring high safety and privacy~\cite{zugner2018nettack}. 

To ensure robust graph learning, many previous efforts are devoted to refining corrupted node features or identifying problematic edges. According to the homophily assumption, AirGNN~\cite{liu2021AirGNN} designs an adaptive message passing scheme with residual connection to boost the resilience to abnormal features. MAGNET~\cite{zhou2023magnet} automatically detects locally corrupted feature attributes and reconstructs robust estimations with sparsity promoting regularizer. Meanwhile, structure learning methods, such as ProGNN~\cite{jin202ProGNN} and GSR~\cite{zhao2023GSR}, aim to learn an optimal graph structure from noisy input and thus alleviate the structure noise or attacks.

However, most of these approaches assume a single noise source of edge perturbation or feature poisoning, purify the graph structures based on feature similarity, or refine node features based on reliable structures. They would face heavy performance degradation when feature noise and structure perturbation exist \textit{simultaneously}. In addition, we find that simply combining the two processes into a pipeline cannot achieve a satisfactory performance and sometimes even gets worse (detail at Sec~\ref{sec:numcomp}). Therefore, reconciling this dilemma remains an unexplored challenge.

To unravel the deadlock, we present a \textbf{U}nified \textbf{G}raph \textbf{D}enoising framework (termed as UGD) that considers both structure and feature noises. Specifically, we propose to conduct structure denoising by developing high-order neighborhood proximity instead of computing the directed similarity of pair nodes, which may be easily disturbed by feature noise. For feature denoising, we leverage a graph auto-encoder to reconstruct node features guided by local smoothness. Furthermore, an iterative updating algorithm is adopted to infer the preserved edges and jointly optimize the node features step by step.

We conduct extensive experiments on various datasets to evaluate the proposed UGD framework. The results validate its superior performance with noise and poisoning attacks over state-of-the-art methods. In particular, numerical comparison on downstream node classification tasks shows that UGD improves accuracy by 4.0\% on the Citeseer dataset and 6.7\% on the AComp dataset over the best previous methods, respectively. Moreover, detailed analyses under different ratios of perturbations and visualizations are also provided to show the strength and advantages of our UGD.

\section{The Proposed UGD Framework}

\subsection{Problem Formulation}
Let $\mathcal{G=(V,E)}$ be an graph with node set $\mathcal{V}$ and edge set $\mathcal{E}$. The feature matrix of the graph is $X \in \mathbb{R}^{n \times d}$, where $n = |\mathcal{V}|$ denotes the number of nodes and $d$ denotes the number of input features. $A \in \{0,1\}^{n \times n}$ is the adjacency matrix, where $A_{uv}=1$ means the node $u$ and node $v$ are connected. $N(u)$ is the direct neighbors of node $u$.
A graph data can also be denoted as $\mathcal{G} = \{\mathcal{E},X\}$. 

\begin{figure}[tbp]
\begin{adjustbox}{center}
    \centering
    \includegraphics[width=1.1\linewidth]{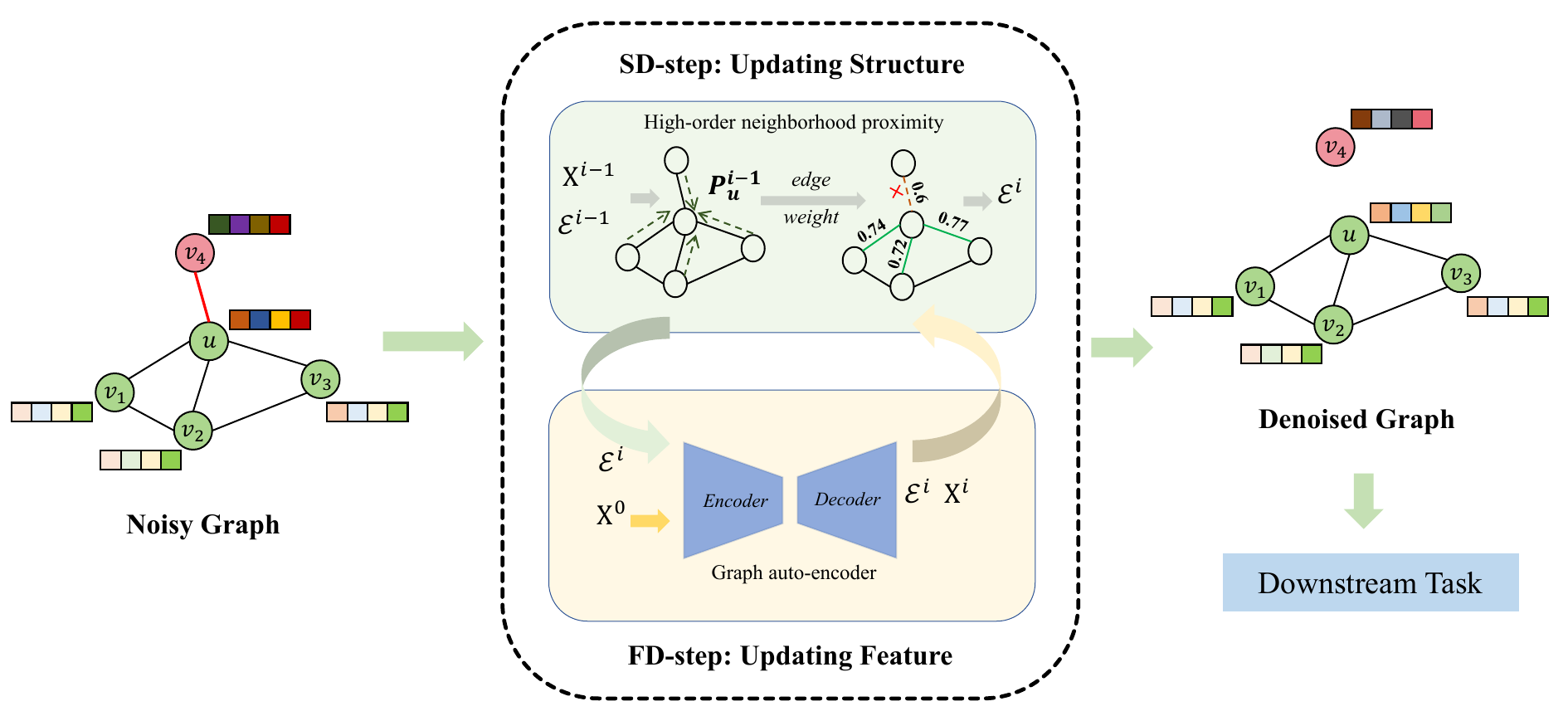}
\end{adjustbox}
\caption{The overall framework of UGD. Node $u$ is injected with feature noise and the edge $(u,v_4)$ is an adversarial link.}
\label{fig:arch}
\end{figure}

Given a noisy or poisoned graph $\mathcal{G} = \{\mathcal{E}, X\}$, the unified denoising problem in this paper is to produce an optimized graph $\mathcal{G^*} = \{\mathcal{E}^*, X^*\}$ considering both structure and feature, and $\mathcal{G^*}$ will be used for the following downstream graph learning tasks. The overall framework of UGD is illustrated in Figure~\ref{fig:arch}.

\subsection{High-order Neighborhood Proximity}\label{sec:structure denoise}

Most existing works in graph structure learning focus on direct relations between interconnected nodes, with metrics like similarity measures or attention mechanisms to remove anomalous edges. However, in the presence of nodes with high feature noise, these pairwise metrics may become less reliable, potentially leading to the erroneous removal of normal edges.  
An intuitive way is to make a node observe more neighborhood information and keep up with more friends. From the perspective of a central node, its identity can be better represented by the neighbors and thus be resistant to feature disturbance. Thus, we define the neighborhood prototype of node $u$ as an aggregating vector $P_u$ of its neighbors with a permutation-invariant readout function. For simplicity, we apply the mean pooling as $P_u = \sum_{v \in N(u)}\frac{x_v}{|N(u)|}$.

Subsequently, for each edge $(u, v)$ in the noisy graph $\mathcal{G}$, we introduce a function $sim(\cdot,\cdot)$ to measure high-order neighborhood proximity $ D_{(u,v)}$ between node $v$ and prototype $P_u$, and distinguish whether the interaction of edge $(u, v)$ is positive or negative. 
In this work, the cosine similarity score $D_{(u,v)}=\frac{P_u\cdot x_v}{\|P_u\|_{2} \|x_v\|_{2}}$ is adopted for its better performance. Considering the asymmetry of $D_{(u,v)}$ and $D_{(v,u)}$ in undirected graphs, the smaller of the two values is retained as weight to filter more noise and formulated as:
\begin{align}
    weight_{(u,v)} = weight_{(v,u)}= min \{D_{(u,v)},D_{(v,u)}\},
\end{align}
where $weight_{(u,v)}$ goes beyond simple consideration of relations between adjacent nodes, incorporating the features of second-order neighbors for a more robust judgment of noisy edges. A threshold $\theta$ is employed to selectively remove edges with lower weights, thereby preserving a credible graph structure.

\subsection{Feature Reconstruction}\label{sec:feature denoise}
To alleviate the impact of abnormal features and achieve local smoothness, prior approaches~\cite{liu2021AirGNN,liu2021elastic} based on graph signal denoising often formulate it as a complex convex optimization problem. Nevertheless, graph anomaly detection methods, such as CONAD~\cite{xu2022conad}, have observed that feature compression and reconstruction can effectively identify anomalous nodes.  
Motivated by this, we develop a graph auto-encoder to ensure that the reconstructed features deviate from the original noise while maintaining similarity with its neighbors. In addition, we introduce a residual connection, combining the reconstructed results with the original features in a weighted manner to form the denoised features. With a fundamental reconstruction loss function $\mathcal{L}_{recon}$, the forward propagation process of node feature $X$ can be defined as follows:
\begin{gather}
    \hat{X}=\beta \cdot X + (1-\beta) \cdot Dec(Enc(X |\mathcal{E} ) |\mathcal{E} ), \\
    \mathcal{L}_{recon}=|| \hat{X}-X||_{2} = \frac{1}{| \mathcal{V}|}\sum_{v \in \mathcal{V}} || \hat{x}_v - x_v ||_{2},
\end{gather}
where the encoder $Enc$ and decoder $Dec$ are two-layer GCNs~\cite{kipf2016semi}, $\hat{X}$ is the denoised feature and $\beta$ is a residual weight parameter.

To better achieve local smoothness in the reconstructed features, we also introduce a neighborhood smoothness loss ($\mathcal{L}_{smooth}$) to enhance the representational capacity of denoised features. Formally, the objective is to force a neighborhood proximity:
\begin{align}
    \mathcal{L}_{smooth}=\frac{1}{2} \sum_{v \in \mathcal{V}}\sum_{u\in N(v)} ||\frac{\hat{x_u}}{\sqrt{d_u}}-\frac{\hat{x_v}}{\sqrt{d_v}}||_2^2 = tr(\hat{X}^TL\hat{X}),
\end{align}
where $L$ is the normalized Laplacian matrix derived on the structure of graph $\mathcal{G}$, $d_v$ is the degree of node $v$. 
The reconstructed feature $\hat{X}$ can be optimized by minimizing the two losses:
\begin{align}
    \mathcal{L}=\mathcal{L}_{recon} (\hat{X} | X)+\gamma \cdot \mathcal{L}_{smooth}(\hat{X} | \mathcal{E}, X) ,
\end{align}
with $\gamma$ being a balance weight of the reconstruction and smoothness.

\subsection{Optimization with Iterative Updating}\label{sec:em}
Since the graph structure and node features are interrelated, the overall objective function $\mathcal{L}$ is hard to optimize. To address this, we build the optimization of structure denosing (SD) and feature denoising (FD) with an iterative updating (IU) algorithm.

In each iteration $i$, the SD-step aim to find a credible edge subset while minimizing the ratio of noisy edges. At this point, node features are fixed, and the edges are updated by our defined high-order neighborhood proximity in section~\ref{sec:structure denoise}:
\begin{align}
    update \ \  \mathcal{E}^i= \{ (u,v) | \ (u,v) \in \mathcal{E}^{i-1} and \ weight_{(u,v)}^{i-1} \geq \theta \} ,
\end{align}
where $\mathcal{E}^i$ is the updated structure.

Then the target is redirected to update node feature in FD-Step. Model parameters of the graph auto-encoder are optimized with the objective function in section~\ref{sec:feature denoise}:
\begin{align}
    minimize \  \mathcal{L}= \mathcal{L}_{recon}(X^i | X^0) + \gamma \cdot \mathcal{L}_{smooth}(X^i | \mathcal{E}^i, X^0).
\end{align}
These two steps would alternate until the estimation $\mathcal{E}^i$ achieves convergence that the difference between two iterations is not more than a stop number $\epsilon$. 

With the designs above, we build our proposed UGD framework and purify the graph data to support subsequent learning of downstream tasks. 
UGD is self-supervised and can be easily implemented as a plug-and-play module for various models and tasks.

\section{Experiments}

\subsection{Experimental Setup}\label{sec:baseline}
\noindent\textbf{Datasets.} We evaluate the semi-supervised node classification performance of UGD on five widely used benchmark datasets~\cite{kipf2016semi,shchur2018pitfalls}. The dataset statistics are summarized in Table \ref{tab:datasets}.

\noindent\textbf{Baselines.} We compare the proposed UGD with representative GNNs and defense models, including: 
    \textbf{(1) Vanilla GCN}~\cite{kipf2016semi} that ignores noises.
    \textbf{(2) Feature denoising methods}: MAGNET~\cite{zhou2023magnet} uses a sparsity promoting regularizer to recover robust estimations of features; AirGNN~\cite{liu2021AirGNN} and ElasticGNN~\cite{liu2021elastic} focus on local smoothness by combining $l_1$ and $l_2$ loss.
    \textbf{(3) Structure learning methods}: Pro-GNN~\cite{jin202ProGNN} directly takes the adjacency matrix as parameters for learning; IDGL~\cite{chen2020IDGL} and GSR~\cite{zhao2023GSR} utilize metric learning to refine the structure.
    \textbf{(4) Combination} of feature denoising and structure learning methods, including different models and orders.

\noindent\textbf{Settings.}
For a fair comparison, we closely follow the experiment setting in previous works~\cite{zhou2023magnet,jin202ProGNN}. We employ \textit{attribute injection} and \textit{metattack}~\cite{zügner2018adversarial} to inject the feature noise and modify the graph structure, respectively. For UGD, the optimal threshold $\theta$ for structure denoising varies across different datasets. We employ a smaller threshold during the early iterations to prevent removing excessive edges due to noisy features. Other hyper-parameters are tuned from: (1) $\beta$: \{0, 0.5\}; (2) $\gamma$: \{1e-5, 5e-4, 3e-4, 1e-3\}; (3) learning rate $\eta$: \{5e-4, 1e-3\};  (4) stop number $\epsilon$: \{0, 2, 10, 30\};
The backbone model for downstream node classification is a two-layer GCN~\cite{kipf2016semi} for all methods, with a learning rate of 0.01, weight decay of 1e-3, and trained for 100 epochs. Adam optimizer~\cite{kingma2014adam} is used for all experiments. Our methods are implemented using PyTorch and PyTorch Geometric~\cite{Fey/Lenssen/2019pyg}. All experiments are conducted on a machine with an NVIDIA 3090 GPU (24GB memory).

\begin{table}[t]
    \centering
    \caption{Statistics of the benchmark graphs.}
    \label{tab:datasets}
    \renewcommand\arraystretch{1}
    \small
    \scalebox{0.95}{
    \begin{tabular}{lccccc}
    \toprule
    \textbf{Dataset} & \#Nodes & \#Edges  & \#Features & \#Classes & \#Train/Val/Test \\
    \midrule
    Cora & 2708 & 5278 & 1433 & 7 & 140/500/1000 \\
    Citeseer & 3327 & 4552 & 3703 & 6 &120/500/1000 \\
    Pubmed & 19717 & 44324 & 500 & 3 & 60/500/1000\\
    AComp & 13752& 245861& 767 & 10 & 200/300/Rest\\
    Coauthor &18333 & 81894 & 6805 & 15 & 300/450/Rest\\
    \bottomrule
    \end{tabular}
    }
\end{table}

\begin{table}[t]
    \caption{Mean accuracy ± stdev over benchmark graphs.}
    \centering
    \label{tab:experiment}
    \renewcommand\arraystretch{1}
    \small
    \scalebox{0.95}{
        \begin{tabular}{l|ccccc}
        \toprule
        \textbf{Methods} & Cora & Citeseer & Pubmed & AComp & Coauthor  \\
        \midrule
        GCN~\cite{kipf2016semi}         &56.5\tiny$\pm$2.0 &42.1\tiny$\pm$2.9 & 65.5\tiny$\pm$1.1 & \underline{68.4\tiny$\pm$2.7}& \underline{81.4\tiny$\pm$2.5}\\
        MAGNET~\cite{zhou2023magnet}      & \underline{69.7\tiny$\pm$2.4} &52.6\tiny$\pm$1.9 & OOM   & OOM & OOM\\
        AirGNN~\cite{liu2021AirGNN}      &68.0\tiny$\pm$2.8 &50.1\tiny$\pm$2.2 & \underline{69.1\tiny$\pm$2.9} & 57.1\tiny$\pm$4.3& 80.7\tiny$\pm$2.8\\
        ElasticGNN~\cite{liu2021elastic}  &68.3\tiny$\pm$3.3 &\underline{52.8\tiny$\pm$3.0} & 66.7\tiny$\pm$1.3 & 58.0\tiny$\pm$4.3& 75.7\tiny$\pm$2.9\\
        \midrule
        Pro-GNN~\cite{jin202ProGNN}     &62.9\tiny$\pm$3.1 &44.8\tiny$\pm$1.0  & 65.6\tiny$\pm$2.3 & 58.8\tiny\tiny$\pm$2.6 & 80.7\tiny$\pm$3.1 \\
        IDGL~\cite{chen2020IDGL}        &55.2\tiny$\pm$2.4  &37.3\tiny$\pm$2.4  & 65.1\tiny$\pm$0.7  & 37.3\tiny$\pm$17.5 & OOM\\
        GSR~\cite{zhao2023GSR}         &58.5\tiny$\pm$3.8  &41.5\tiny$\pm$1.0  & 64.4\tiny$\pm$1.0  & 26.1\tiny$\pm$8.0 & 77.3\tiny$\pm$6.3 \\
        \midrule
        MAGNET+GSR  & 69.4\tiny$\pm$2.4   & 51.6\tiny$\pm$1.5  & OOM & OOM& OOM\\
        AirGNN+GSR  & 57.1\tiny$\pm$2.8   & 40.2\tiny$\pm$2.1  & 66.3\tiny$\pm$1.7  &57.0\tiny$\pm$3.5  &77.1\tiny$\pm$4.1  \\
        MAGNET+IDGL    &67.5\tiny$\pm$2.8  &50.7\tiny$\pm$1.9  &OOM & OOM& OOM\\
        GSR+MAGNET     &68.6\tiny$\pm$2.2  &51.8\tiny$\pm$2.2  &OOM & OOM& OOM\\
        GSR+AirGNN     &69.0\tiny$\pm$1.9  &50.1\tiny$\pm$2.6 &67.2\tiny$\pm$2.8  &52.2\tiny$\pm$7.8  &78.4\tiny$\pm$3.8 \\
        \midrule
        UGD (ours)          &\textbf{70.3\tiny$\pm$1.2 } & \textbf{54.9\tiny$\pm$3.2 } & \textbf{70.0\tiny$\pm$1.4 } & 
        \textbf{73.0\tiny$\pm$1.3 } &\textbf{83.2\tiny$\pm$1.8 }  \\
        \midrule
        \textit{w/o HNP}       &69.7\tiny$\pm$2.8 &54.4\tiny$\pm$4.4 & 69.4\tiny$\pm$1.8 & 72.8\tiny$\pm$1.5& 83.1\tiny$\pm$2.6\\
        \textit{w/o FR}       &58.4\tiny$\pm$2.1  &43.0\tiny$\pm$2.5  & 66.1\tiny$\pm$1.5  & 69.3\tiny$\pm$2.3 & 81.6\tiny$\pm$2.3 \\
        \textit{w/o IU (F+S)}      &69.7\tiny$\pm$2.3  & 54.8\tiny$\pm$3.8& 69.3\tiny$\pm$1.2  & 72.4\tiny$\pm$1.5 & 82.9\tiny$\pm$2.4 \\
        \textit{w/o IU (S+F)}  &68.9\tiny$\pm$2.5&53.7\tiny$\pm$3.1&69.0\tiny$\pm$2.1&72.2\tiny$\pm$2.3&83.0\tiny$\pm$2.3 \\
        \bottomrule
        \end{tabular}
        }
\end{table}

\subsection{Numerical Comparison}
\label{sec:numcomp}
We evaluate the performance of all methods against two types of noise, with an injection of 10\% structure noise and 50\% feature noise into the data. The ratio settings are referenced from prior works~\cite{zhou2023magnet,liu2021elastic}, but we are pioneering in simultaneously considering both of them. Tabel~\ref{tab:experiment} summarizes the results of downstream node classification tasks after denoising. We report the mean test accuracy, together with a standard deviation over five different random seeds (OOM means out of memory). We can make the following observations from the results:

\begin{figure*}[tbp]
  \centering
  \begin{subfigure}[b]{0.3\textwidth}
    \includegraphics[width=1\textwidth]{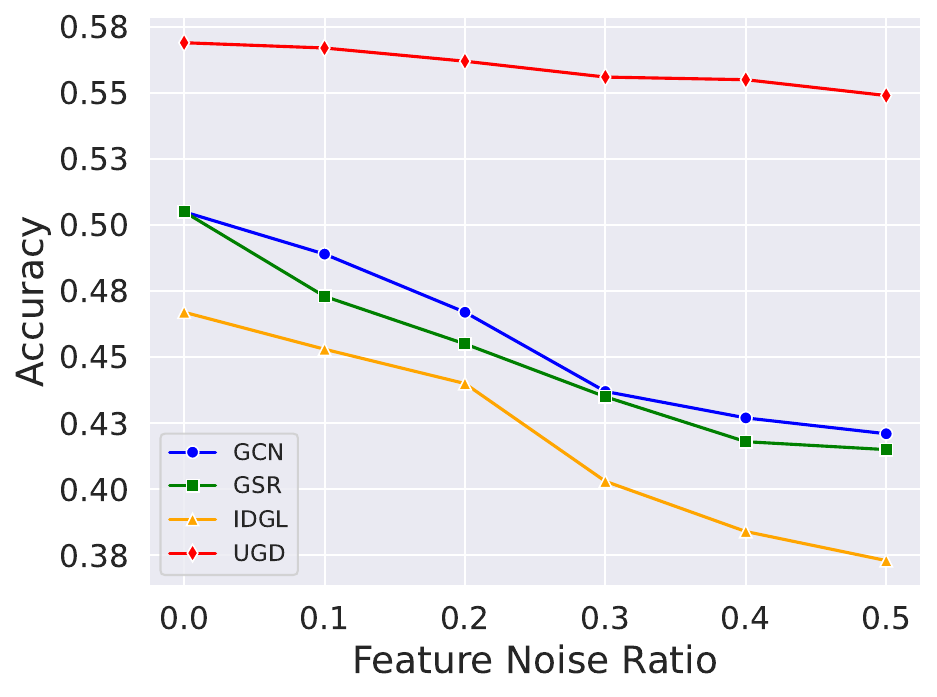}
    \caption{Fixed structure noise}
    \label{fig:sub1}
  \end{subfigure}
  \hfill
  \begin{subfigure}[b]{0.3\textwidth}
    \includegraphics[width=1\textwidth]{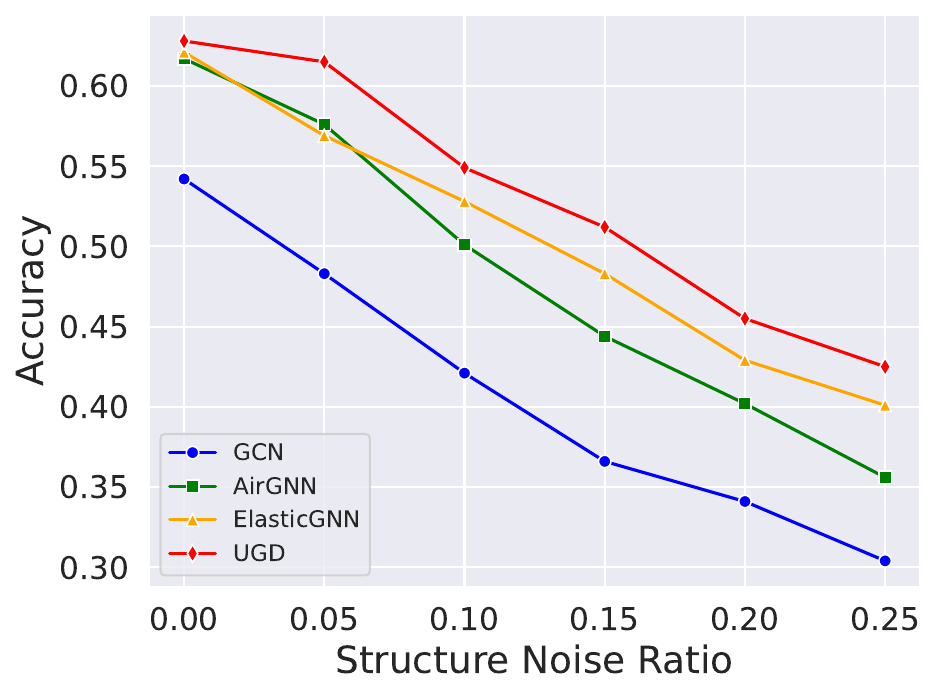}
    \caption{Fixed feature noise}
    \label{fig:sub2}
  \end{subfigure}
  \hfill
  \begin{subfigure}[b]{0.38\textwidth}
    \includegraphics[width=1\textwidth]{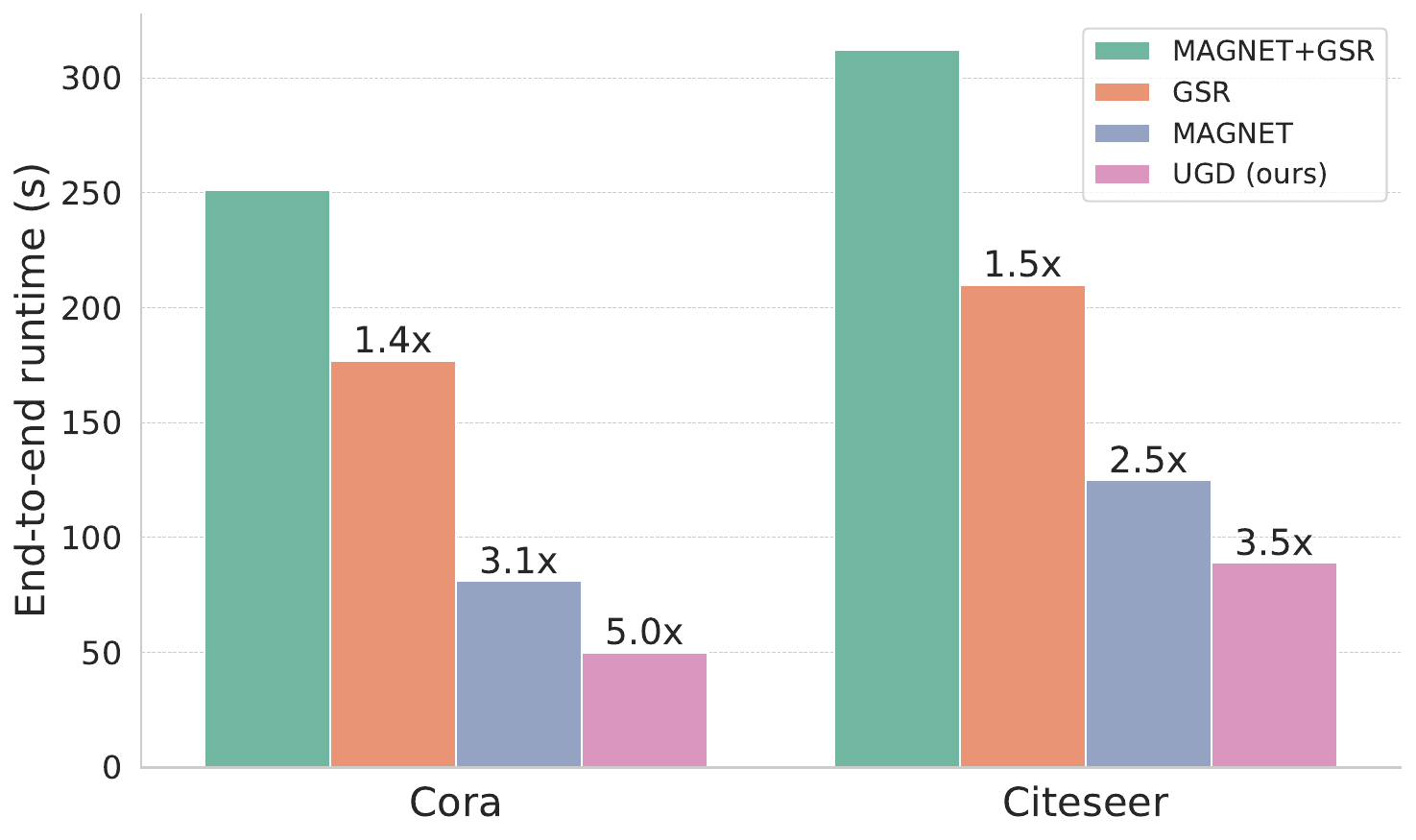}
    \caption{Efficiency comparison}
    \label{fig:sub3}
  \end{subfigure}
  \caption{Model analysis.}
  \label{fig:feature}
\end{figure*}

    (1) The performance of feature denoising methods can be heavily affected by structure noise. Specifically, on the AComp and Coauthor datasets, the elaborate approaches perform even worse than vanilla GCN. Besides, structure learning methods have shown large fluctuations in performance with feature noise. GSR performs much worse than GCN on most datasets (except for Cora). These results demonstrate the necessity of considering both types of noise.
    
    (2) Simply connecting the methods cannot achieve a satisfying result. For example, AirGNN+GSR leads to performance degradation compared with a single AirGNN on all graphs, which confirms our motivation to develop a unified denoising framework.

    (3) With the iterative updating algorithm, UGD jointly optimizes structure and feature denoising, and significantly improves the performance of GNNs. Concretely, UGD consistently outperforms the best baselines across five datasets, obtaining improvements of 0.9\%, 4.0\%, 1.3\%, 6.7\%, and 2.2\%, respectively.

\subsection{Ablation Study}
To examine the contributions of different components in UGD, we also consider variants of UGD that serve as ablation: (1)\textit{w/o HNP}, which focuses solely on features denoising without refining the graph structure, and (2)\textit{w/o FR}, which applies structure denoising without feature reconstruction. (3)\textit{w/o IU}, which removes the iterative updating algorithm and performs graph denoising in order of feature first (\textit{F+S}) or structure first (\textit{S+F}). 

In Table~\ref{tab:experiment}, the bottom part summarizes the results, from which we have two observations. First, the performance of all UGD variants drops distinctly when compared to the full model, demonstrating that each designed component contributes to the success of UGD. Second, UGD without high-order proximity still outperforms most feature denoising methods, and UGD without feature reconstruction is still comparable to most structure learning methods, verifying the effectiveness of our designs under mixed noises.

\subsection{Model Analysis}
We have performed comprehensive analysis to verify the advantages and effectiveness of UGD. Limited by the page, we report main results on Citeseer as following:

\subsubsection{Performance on different feature noise ratios.}
Maintaining a consistent structure noise ratio of 10\%, we conduct experiments with varying feature noise ratios of 0\% to 50\% on Citeseer dataset. The experimental results are presented in Figure~\ref{fig:feature}(a). In this study, we compare our UGD model with GCN and other GSL methods to analyze how feature noise affects the structure denoising capability of the models. It can be observed that when the feature noise ratio increases, the baselines' node classification accuracy drops fast. UGD consistently outperforms baselines with a significant margin. Concretely, the accuracy on Citeseer drops by 18\% and 21\% for GSR and IDGL, respectively. For UGD, the accuracy decreases from 0.57 to 0.55, resulting in a slight decay of 3.5\%, indicating that UGD is more robust against feature noise.

\subsubsection{Performance on different structure noise ratios.} Maintaining a consistent feature noise ratio of 50\%, we also compare UGD with graph denoising methods with different ratios of structure noise varying from 0\% to 25\% on Citeseer dataset. As shown in Fig.~\ref{fig:feature}(b), structure noise can severely affect the graph denoising methods. Nevertheless, our UGD's performance is superior to baselines in most cases, showing better robustness against structure noise.

\subsubsection{Efficiency.}
Another advantage of UGD is its high efficiency in graph denoising, which could be essential for large graphs. We report the end-to-end runtime of denoising methods with the same epochs in Fig.\ref{fig:feature}(c). 
Compared to MAGNET+GSR, UGD can achieve a speedup of 5.0x and 3.5x on Cora and Citeseer, respectively. In fact, UGD performs faster than single MAGNET and GSR, showing potential in applications with high-efficiency requirements. Moreover, UGD is model-free and can be readily combined with scalable GNNs to handle large graphs.

\section{Conclusion}
In this paper, we focus on robust graph learning with challenging structure and feature perturbations and introduce a UGD framework to handle structure and feature denoising for downstream tasks simultaneously. In UGD, high-order neighborhood proximity and feature reconstruction are jointly optimized with an iterative updating algorithm. Experiments on various datasets demonstrate the significant effectiveness of UGD. Extensive analyses further verify UGD's advantages against different noise and high efficiency.


\begin{acks}
This work was supported by National Key R\&D program of China under Grant No.2023YFC3807603.
\end{acks}

\bibliographystyle{ACM-Reference-Format}
\balance
\bibliography{sample-base}




\end{document}